\documentclass{article}

\PassOptionsToPackage{numbers,compress}{natbib}

\usepackage[preprint]{neurips_2026}

\usepackage[utf8]{inputenc} % allow utf-8 input
\usepackage[T1]{fontenc}    % use 8-bit T1 fonts
\usepackage{hyperref}       % hyperlinks
\usepackage{url}            % simple URL typesetting
\usepackage{booktabs}       % professional-quality tables
\usepackage{amsfonts}       % blackboard math symbols
\usepackage{microtype}      % microtypography
\usepackage{xcolor}         % colors

\usepackage{float}
\usepackage{amsmath,amssymb}
\usepackage{graphicx}
\usepackage{xspace}
\usepackage{tabularx}
\usepackage{array}
\usepackage{tikz}
\usetikzlibrary{arrows.meta,positioning,fit,calc,decorations.pathreplacing}

\usepackage{todonotes}

\definecolor{refusalblue}{HTML}{3B6FB6}
\definecolor{sycorange}{HTML}{D7792D}
\definecolor{naturalgreen}{HTML}{3A8D6D}
\definecolor{mutedgray}{HTML}{707070}
\definecolor{lightblue}{HTML}{DCE8F5}
\definecolor{lightorange}{HTML}{F5E4D5}
\definecolor{lightgreen}{HTML}{DCEEE7}

\newcommand{\model}{Qwen2.5-7B-Instruct\xspace}
\newcommand{\faith}{\mathrm{F}}
\newcommand{\necess}{\mathrm{N}}

\title{From Trait Vectors to Circuits: Tracing Refusal and Sycophancy Through Language Models}

\author{%
  \begin{tabular}{@{}>{\centering\arraybackslash}p{0.44\textwidth}@{\hspace{1em}}>{\centering\arraybackslash}p{0.44\textwidth}@{}}
    Oscar Miró López-Feliu\textsuperscript{1}
    &
    Maya Ozbayoglu\textsuperscript{1,2} \\[2pt]
    {\normalfont\small\texttt{oscar.miro.lopez.feliu@student.uva.nl}}
    &
    {\normalfont\small\texttt{maya@safeainetherlands.org}} \\[3pt]
    \multicolumn{2}{c}{\normalfont\small
      \textsuperscript{1}\,University of Amsterdam
      \qquad
      \textsuperscript{2}\,Safe AI Netherlands}
  \end{tabular}
}

\begin{document}

\maketitle

\begin{abstract}
A direction in activation space that changes safety-relevant behavior when steered is not necessarily one the model uses to produce that behavior on its own. We therefore ask whether steering acts through the computation of the unmodified model or through a different set of components, studying two traits whose directions have been extracted and validated in prior work: refusal and sycophancy in Qwen2.5-7B-Instruct. For each, we use the trait vector to split the computation into a reconstruction circuit before the vector and a transmission circuit after it. We then test whether restoring the coordinate returns behavior removed by ablation. For refusal, both circuits are compact and faithful, restoring the coordinate alone recovers almost all of the refusal signal lost to ablation, and a circuit built around the vector matches the faithfulness of a direct input-to-output circuit at roughly half the edges. For sycophancy, transmission is compact, but reconstruction is broader and only partially faithful. Circuits that transmit a steering intervention are therefore not automatically the circuits that produce the behavior, and we report the counterfactual, target, and behavioral test for every circuit accordingly.
\end{abstract}

\section{Introduction}

% is this to be deleted? was deleted but not sure if that is an error bc we do need a hook cant start a paper with prior work
Determining whether an intervention acts through the computation used by the unmodified model or controls behavior through a different pathway is central to understanding how language models implement safety-relevant behaviors. 
Prior work identifies linear directions in model activations for refusal \citep{arditi2024refusal}, truth \citep{marks2024the}, sentiment \citep{tigges2024language}, and persona traits \citep{chen2025persona}. Adding or removing these directions can change model behavior \citep{braun2025understanding, rimsky-etal-2024-steering, arditi2024refusal, chen2025persona}, although steering effects vary across prompts and traits \citep{braun2025understanding}. While this provides researchers with a useful intervention technique, it does not explain the complete computation process that produces the behavior. 

Consider a refusal direction measured at an intermediate layer. Components before this layer contribute to the model state's projection onto the direction, while components after it transform the state into token predictions across a generated refusal. Steering may act through the same computation used by the unmodified model or through a different set of components. A circuit that transmits the effect of steering therefore does not necessarily explain how the model normally produces refusal.

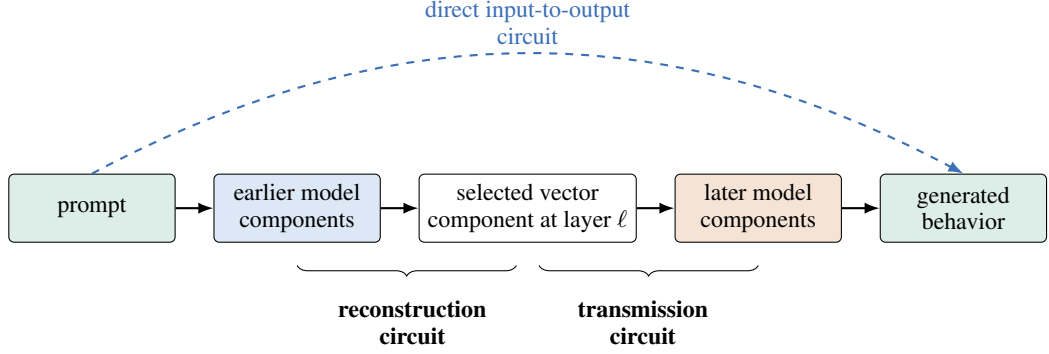
\begin{figure}[t]
\centering
\begin{tikzpicture}[
  node distance=5mm,
  box/.style={draw, rounded corners=2pt, minimum height=9mm, minimum width=22mm, align=center, font=\small},
  arr/.style={-{Latex[length=2mm]}, thick},
  tag/.style={font=\small\bfseries, align=center}
]
\node[box, fill=lightgreen] (input) {prompt};
\node[box, fill=lightblue, right=of input] (early) {earlier model\\components};
\node[box, fill=white, right=of early] (vector) {selected vector\\component at layer $\ell$};
\node[box, fill=lightorange, right=of vector] (late) {later model\\components};
\node[box, fill=lightgreen, right=of late] (output) {generated\\behavior};
\draw[arr] (input) -- (early);
\draw[arr] (early) -- (vector);
\draw[arr] (vector) -- (late);
\draw[arr] (late) -- (output);
\node[tag, below=11mm of $(early)!0.5!(vector)$] {reconstruction\\circuit};
\node[tag, below=11mm of $(vector)!0.5!(late)$] {transmission\\circuit};
\draw[decorate, decoration={brace,mirror,amplitude=4pt}] ([yshift=-3mm]early.south) -- ([xshift=-1.5mm,yshift=-3mm]vector.south);
\draw[decorate, decoration={brace,mirror,amplitude=4pt}] ([xshift=1.5mm,yshift=-3mm]vector.south) -- ([yshift=-3mm]late.south);
\draw[arr, dashed, refusalblue, bend left=28] (input.north) to node[above,align=center,font=\small] {direct input-to-output\\circuit} (output.north);
\end{tikzpicture}
\vspace{-0.7em}
\caption{We divide computation around a selected trait vector. Under a specified corruption, the reconstruction circuit recovers the vector component from an earlier source. The transmission circuit carries a source-local vector intervention to model outputs. A direct input-to-output circuit provides a comparison.}
\label{fig:overview}
\vspace{-1.0em}
\end{figure}

We use a known trait vector at a selected layer to define two circuit-discovery problems. A \emph{reconstruction circuit} identifies edges that recover the activation component along the selected vector under a stated corruption or paired replacement. A \emph{transmission circuit} identifies later edges that carry a source-local vector intervention into generated-token predictions. We use EAP-IG \citep{hanna2024have} for both problems, then test exact restoration of this vector component, whether the transmission circuit carries its effect, and direct input-to-output circuits. Figure~\ref{fig:overview} summarizes the reconstruction and transmission circuits and the direct input-to-output comparison.

We study refusal and sycophancy in \model. For refusal, we use the prompt-derived direction of \citet{arditi2024refusal}. For sycophancy, we study both a prompt-derived direction and the response-average persona direction of \citet{chen2025persona}. We ask:
\begin{enumerate}
    \item Does changing the activation component along the selected trait vector causally change behavior?
    \item Which edges reconstruct or support this vector component under a stated counterfactual?
    \item Which later edges transmit a source-local vector intervention into multi-token behavior?
    \item How much of each tested effect is mediated by the selected coordinate?
    \item How does this vector-based decomposition compare with direct input-to-output circuit discovery under a matched target?
\end{enumerate}

Our main contributions are:
\begin{itemize}
    \item We separate five questions: whether a trait vector controls behavior, which earlier edges reconstruct its activation component, which later edges transmit its effect, whether restoring that component reverses a circuit intervention, and how this decomposition compares with direct input-to-output circuits.
    \item For refusal, a compact reconstruction subgraph recovers the selected vector component, and a separate transmission subgraph carries recurrent steering. Restoring the refusal-vector component after reconstruction-subgraph ablation recovers most of the prompt-end effect. Under a matched static target, a two-stage construction reaches comparable faithfulness with fewer edges and preserves more refusal than an equal-budget direct circuit.
    \item For sycophancy, prompt and response interventions have complementary behavioral effects, and a compact transmission subgraph carries response-vector steering. In contrast, input-connected reconstruction is substantially broader, and replacing only the response-direction component explains less than one-third of the tested natural behavioral change.
\end{itemize}

\section{Related Work}

\paragraph{Trait vectors and activation steering.}
Activation addition, contrastive activation addition, representation engineering, and inference-time intervention use directions in activation space to control model behavior \citep{turner2023activation,rimsky-etal-2024-steering,zou2023representation,li2023inferencetime}. \citet{arditi2024refusal} extract refusal directions from harmful and harmless prompts and show that adding or removing them changes refusal. \citet{chen2025persona} primarily derive persona directions by averaging response-token activations, while also comparing prompt-last and prompt-average alternatives. Together, these studies motivate our analysis of how earlier components reconstruct these directions and how later components carry their effects to model outputs.

\paragraph{Circuit discovery.}
Path patching, automated circuit discovery, and edge attribution patching identify components and edges that contribute to model behavior \citep{goldowskydill2023path,conmy2023towards,syed-etal-2024-attribution}. EAP-IG estimates each edge's contribution by averaging gradients along a linear path between corrupted and clean activations \citep{hanna2024have}. Prior work shows that the corruption, target metric, and patching direction can change which components appear important \citep{zhang2023best,heimersheim2024patching}. Related causal-mediation methods learn sparse activation dimensions or select model components that mediate long-form behavior \citep{doan-etal-2026-causal,sankaranarayanan2025activation}. We use EAP-IG to identify circuits before and after a selected trait vector, then separately compare circuits defined by vector interventions with circuits defined by natural input contrasts.

\paragraph{Refusal and sycophancy mechanisms.}
Recent work complicates a single-direction account of refusal by identifying multiple refusal directions and concept cones, separating harmfulness detection from refusal, and finding sparse refusal-related features \citep{wollschlager2025the,hawasly2026more,zhao2025llms,obrien2025steering}. \citet{cheng2026drives} use multi-token activation patching to identify circuits that carry refusal steering through generated responses. Our transmission experiments use their tokenwise comparison of steered and unsteered predictions. We additionally identify earlier edges that reconstruct the refusal vector, test whether restoring its activation component reverses reconstruction-circuit ablation, and compare this two-stage description with direct input-to-output circuit discovery. Prior work documents sycophancy in free-form generation and identifies late-layer preference shifts, influential attention heads, and circuits shared with factual or instructed lying \citep{sharma2024towards,wang2025truth,genadi-etal-2026-sycophancy,pandey2026llms}. We study sycophancy induced by paired sycophantic and non-sycophantic instructions and test whether its response-average persona direction also explains the behavioral difference between them.

\section{Method}
\subsection{Causal questions around a trait vector}
Let \(h_\ell(x,t)\) be the residual state at layer \(\ell\), token \(t\), and input \(x\). For a unit direction \(v\), the scalar coordinate is \(z_v(x,t)=v^\top h_\ell(x,t)\). We begin with an intervention-only baseline. A circuit that carries a vector intervention is behaviorally meaningful only if changing the vector changes behavior, and the full intervention defines the effect that the transmission circuit should reproduce. We use five tests corresponding to the five questions above:
\begin{enumerate}
    \item We change \(z_v\) and measure the resulting behavior. This establishes whether the vector is causally effective and provides the baseline for later circuit tests.
    \item We find edges before the selected layer that reconstruct a coordinate difference created by a specified corruption or source intervention. This identifies a reconstruction circuit for that comparison, not necessarily the unrestricted process that forms the vector from the input.
    \item We find edges after the selected layer that carry a behaviorally effective vector intervention to generated-token predictions. Circuit-only and circuit-ablation tests measure how much of the full intervention effect passes through these edges.
    \item We test coordinate mediation using graph ablation and restoration for refusal and paired state replacement for sycophancy, in each case alongside a full-state condition.
    \item We discover a direct input-to-output circuit for the complete chosen contrast. This provides a reference for the two-stage reconstruction and transmission analysis.
\end{enumerate}

Reconstruction and transmission analyze computation before and after the selected vector. Coordinate restoration tests whether the selected component mediates the effect of a specified graph intervention. A high transmission score shows that the graph carries the specified intervention. It does not establish that the same edges implement the behavior when no intervention is applied.

\subsection{Circuit metrics}
For each experiment, EAP-IG integrates edge attributions between a clean endpoint and a corrupted endpoint. Let \(m\) be the metric for an experiment and \(C\) a selected set of edges. A circuit-only run uses clean activations on \(C\) and corrupted activations elsewhere. A circuit-ablated run does the reverse. We report
\begin{equation}
\faith(C)=\frac{m(C\text{-only})-m(\text{corrupt})}{m(\text{clean})-m(\text{corrupt})},
\qquad
\necess(C)=\frac{m(\text{clean})-m(C\text{-ablated})}{m(\text{clean})-m(\text{corrupt})}.
\end{equation}
Following \citet{hanna2024have}, we call a circuit faithful when it recovers at least 85\% of the clean-to-corrupt performance difference. Values above 100\% indicate overshoot and are not clipped. We state whether each circuit is source-connected or top-ranked.

\subsection{Reconstruction targets}
Reconstruction targets measure the activation component along the selected direction at a specified layer and set of token positions. Prompt-direction experiments aggregate this target over all prompt tokens, while response-direction experiments aggregate it over generated positions. For refusal, the clean endpoint is the unmodified harmful-prompt execution and the corrupted endpoint globally projects out the refusal direction. For sycophancy, the corrupted endpoint replaces only the selected coordinate with its value from the matched negative-instruction execution at each relevant position, rather than setting it to zero. Appendix~\ref{app:specifications} gives the exact intervention. Natural-contrast experiments instead change the instruction itself. We restrict each graph to components upstream of the target. Because the resulting circuit depends on these endpoints, we call it a reconstruction circuit rather than assuming that it captures natural formation from the input.

\subsection{Transmission targets}
For transmission, we first generate unsteered and recurrently steered responses to the same prompt. The steered trajectory is the clean endpoint and the unsteered trajectory is the corrupted endpoint. We then evaluate both endpoints on the same response prefixes, comparing at every position the logits of the next tokens they prefer and retaining positions where those tokens differ. This shared-prefix target avoids comparing unrelated tokens after independently generated responses diverge. We restrict the graph to the source layer and later components so that the intervention does not directly modify the computation being identified.

\subsection{Mediation through the selected coordinate}
We test coordinate mediation in two settings. For refusal, we ablate a reconstruction graph and restore either the clean coordinate or the complete clean state. For sycophancy, we replace the coordinate, orthogonal state, or complete state of the positive-instruction execution with values from its paired negative-instruction execution. These tests measure how much of the specified circuit or paired-instruction effect is carried by the selected coordinate.

\section{Experimental setup}

\paragraph{Model and graph.}
We use \model \citep{qwen25} in bfloat16 with TransformerLens hooks \citep{transformerlens}. Graph nodes are attention outputs, MLP outputs, residual sources, and experiment-specific targets. Edges represent additive component contributions.

\paragraph{Data splits.}
Refusal extraction uses 128 harmful and 128 harmless PolyRefuse prompts \citep{wang2025refusal}. Sycophancy uses paired positive and negative instructions from the Persona Vectors data \citep{chen2025persona}, with prompt and response directions estimated from the corresponding activations. Vector validation, edge scoring, and circuit selection use disjoint source questions. The final circuit evaluation uses a fixed subset of the behavioral split and remains disjoint from development data. Appendix~\ref{app:data-specifications} gives the complete accounting.

\paragraph{Behavioral evaluation.}
Static refusal experiments use refusal log-odds \citep{arditi2024refusal}, and free generations use WildGuard refusal labels \citep{3737916.3738177}. We score sycophancy and local coherence separately with GPT-4.1-mini \citep{openai2025gpt41}, using fixed evaluator prompts based on the Persona Vectors protocol \citep{chen2025persona}. Appendix~\ref{app:judge-prompts} gives both prompts. Confidence intervals resample questions and recompute paired condition differences, so the independent sample size is the number of questions rather than the number of generated responses. All final balanced 64-question sycophancy circuit-behavior conditions use the same 48-token generation window.

\paragraph{Trait directions and intervention positions.}
We construct each direction as the normalized difference between mean positive and negative activations. The selected refusal direction is the harmful-minus-harmless prompt direction at layer 15. The selected sycophancy directions are the paired positive-minus-negative final-prompt-position direction at layer 15 and the paired response-average direction at layer 20. We select the refusal location using the causal criteria of \citet{arditi2024refusal} and the sycophancy locations using held-out intervention sweeps. Prompt interventions apply the corresponding direction at every prompt position. In the 300-question sycophancy comparison, each direction moves its coordinate by twice the paired positive-to-negative gap, and the response intervention begins after the first-token prediction. The separate transmission experiments apply their source intervention at every prompt position, the first-token prediction, and every decoded prefix. Held-out tests verify that each intervention changes behavior before circuit discovery.

\section{Refusal}
\subsection{Coordinate reconstruction and scalar restoration}

Removing the refusal direction at layer 15 over every prompt position changes mean refusal log-odds from \(4.943\) to \(-4.064\) on 64 held-out harmful prompts. Global projection over all layers changes it to \(-12.642\). We then score edges from the layer-0 residual state after global projection to the summed layer-15 coordinate over all prompt tokens.

A connected reconstruction graph containing 699 of 260,956 candidate edges, or 0.268\%, reaches \(\faith=98.84\%\) and \(\necess=93.90\%\) on 64 held-out prompts. This graph reconstructs the coordinate after global projection; it is not unrestricted formation from the input.

We separately evaluate the 3,000 highest-scoring edges from the same EAP-IG ranking. This set is not the connected 699-edge graph. In free generation on 64 prompts, refusal is 93.8\% normally, 6.3\% after projection, 95.3\% when only these edges retain clean values, and 20.3\% when they are ablated. Ablating the selected graph changes mean prompt-end refusal log-odds from \(4.943\) to \(-9.734\), whereas ablating three matched random graphs leaves them near baseline at \(4.55\)--\(4.70\). Restoring only the clean layer-15 coordinate at every prompt token returns the selected-circuit condition to \(4.099\), recovering 94.2\% of the loss (95\% CI 90.7--97.6\%). Restoring the complete clean layer-15 state returns it to \(4.943\). This is circuit-conditional scalar restoration for a prompt-end metric, not a claim that all natural refusal is one-dimensional.

\subsection{Recurrent intervention transmission}

We apply negative steering with coefficient \(-2\) at layer 15 to every prompt token, the first-token prediction, and every decoded prefix. The clean endpoint is the steered trajectory and the corrupted endpoint is the unsteered trajectory. Discovery uses 64 successful refusal-to-compliance flips selected from 96 prompts; evaluation uses 64 disjoint prompts.

An 8,000-edge connected graph contains 4.12\% of 194,117 candidate edges and reaches \(\faith=87.05\%\) and \(\necess=99.69\%\). In free generation, refusal decreases from 98.44\% without steering to 3.13\% with full steering. Allowing steering to propagate only through the selected graph yields 9.38\% refusal, preserving 93.4\% of the full effect, whereas a matched random graph yields 98.44\% and transmits no detectable effect. Conversely, ablating the selected graph completely blocks steering, restoring refusal to 98.44\%, whereas ablating the random graph leaves steering effective at 7.81\% refusal. This graph transmits the specified recurrent intervention; it is not identified as the unique circuit used for natural refusal.

\subsection{Two-stage versus direct circuit discovery}

We compare two-stage and direct input-to-output circuit discovery under the same global-projection corruption and prompt-end refusal-logit target. The two-stage method discovers one graph before layer 15 and another after it, connecting them through the layer-15 activations produced by the first graph. It transfers the complete resulting activation state, not only the refusal coordinate. The construction combines the connected 699-edge reconstruction graph with a separately discovered 10,000-edge projection-based downstream graph. This downstream graph is distinct from the 8,000-edge recurrent steering-transmission graph above.

At total edge counts of 2,699, 8,199, 8,699, and 10,699, two-stage faithfulness is 70.49\%, 84.45\%, 84.87\%, and 85.61\%, compared with 54.85\%, 73.63\%, 74.39\%, and 71.10\% for equal-budget direct top-ranked subgraphs. The 10,699-edge two-stage construction reaches \(\necess=99.65\%\). The first source-connected direct graph with comparable faithfulness contains 21,605 edges and reaches \(\faith=86.50\%\) and \(\necess=99.13\%\). In a separate 16-prompt free-generation test of this connected direct graph, refusal is 100\% normally, 0\% after global projection, 75\% when only the graph retains clean values, and 0\% when it is ablated. Thus, direct discovery also produces a behaviorally effective connected graph, but at approximately twice the edge count of the two-stage construction.

\begin{figure}[t]
\centering
\begin{minipage}[t]{0.48\linewidth}
\centering
\textbf{(a) Coordinate restoration}\par
\begin{tikzpicture}[x=0.9cm,y=0.18cm,font=\scriptsize]
\draw[->] (0,-11) -- (0,7);
\draw[->] (0,0) -- (5.7,0);
\foreach \y in {-10,-5,0,5}{\draw(-0.08,\y)--(0,\y)node[left]{\y};}
\foreach \x/\h/\c/\lab in {0.75/4.943/refusalblue/Normal,2.0/-9.734/sycorange/Ablated,3.25/4.099/naturalgreen/Coordinate,4.5/4.943/purple/Full state}{
  \fill[\c] (\x-0.3,0) rectangle (\x+0.3,\h);
  \node[below=2pt,rotate=30,anchor=north east] at (\x,-10.3){\lab};
}
\node[above,font=\scriptsize\bfseries] at (3.25,4.5) {94.2\% restored};
\node[rotate=90] at (-0.62,-2) {refusal log-odds};
\end{tikzpicture}
\end{minipage}\hfill
\begin{minipage}[t]{0.49\linewidth}
\centering
\textbf{(b) Recurrent transmission}\par
\begin{tikzpicture}[x=0.76cm,y=0.031cm,font=\scriptsize]
\draw[->] (0,0) -- (0,108);
\draw[->] (0,0) -- (7.2,0);
\foreach \y in {0,25,50,75,100}{\draw(-0.07,\y)--(0,\y)node[left]{\y};}
\foreach \x/\h/\c/\lab in {0.65/98.44/refusalblue/Natural,1.7/3.13/sycorange/Full steer,2.75/9.38/naturalgreen/Graph only,3.8/98.44/refusalblue/Graph abl.,4.85/98.44/mutedgray/Random only,5.9/7.81/mutedgray/Random abl.}{
  \fill[\c] (\x-0.28,0) rectangle (\x+0.28,\h);
  \node[below=1pt,rotate=38,anchor=north east] at (\x,0){\lab};
}
\node[rotate=90] at (-0.55,54) {refusal (\%)};
\end{tikzpicture}
\end{minipage}

\vspace{0.1em}
\begin{minipage}[t]{0.48\linewidth}
\centering
\textbf{(c) Faithfulness versus edge count}\par
\begin{tikzpicture}[x=0.245cm,y=0.073cm,font=\scriptsize]
\draw[->] (0,50) -- (0,97);
\draw[->] (0,50) -- (23,50);
\foreach \y in {50,60,70,80,90}{\draw(-0.25,\y)--(0,\y)node[left]{\y};}
\foreach \x in {0,5,10,15,20}{\draw(\x,49.2)--(\x,50)node[below]{\x};}
\draw[densely dashed,naturalgreen] (0,85)--(23,85);
\node[anchor=south west,naturalgreen,font=\tiny] at (12.8,85) {85\% criterion};
\draw[refusalblue,thick] plot coordinates {(2.699,70.49) (8.199,84.45) (8.699,84.87) (10.699,85.61)};
\foreach \x/\y in {2.699/70.49,8.199/84.45,8.699/84.87,10.699/85.61}{\fill[refusalblue] (\x,\y) circle (1.8pt);}
\draw[sycorange,thick] plot coordinates {(2.699,54.85) (8.199,73.63) (8.699,74.39) (10.699,71.10)};
\foreach \x/\y in {2.699/54.85,8.199/73.63,8.699/74.39,10.699/71.10}{\fill[sycorange] (\x,\y) circle (1.8pt);}
\filldraw[fill=white,draw=sycorange,thick] (21.605,86.50) circle (2.2pt);
\draw[refusalblue,thick] (0.8,94)--(1.8,94);
\fill[refusalblue] (1.3,94) circle (1.8pt);
\node[anchor=west,font=\tiny] at (2.0,94) {two-stage};
\draw[sycorange,thick] (7.2,94)--(8.2,94);
\fill[sycorange] (7.7,94) circle (1.8pt);
\node[anchor=west,font=\tiny] at (8.4,94) {direct top-ranked};
\draw[sycorange,fill=white,thick] (16.0,94) circle (2.2pt);
\node[anchor=west,font=\tiny] at (16.5,94) {direct connected};
\node at (11.5,44.5) {total edges (\(\times 10^3\))};
\node[rotate=90] at (-2.0,73) {faithfulness (\%)};
\end{tikzpicture}
\end{minipage}\hfill
\begin{minipage}[t]{0.49\linewidth}
\centering
\textbf{(d) Two-stage versus direct behavior}\par
\begin{tikzpicture}[x=0.78cm,y=0.031cm,font=\scriptsize]
\draw[->] (0,0) -- (0,108);
\draw[->] (0,0) -- (7.2,0);
\foreach \y in {0,25,50,75,100}{\draw(-0.07,\y)--(0,\y)node[left]{\y};}
\foreach \x/\h/\lo/\hi/\c/\lab in {
0.65/98.44/95.31/100/refusalblue/Normal,
1.7/31.25/20.31/42.19/mutedgray/Projection,
2.75/90.63/82.81/96.88/naturalgreen/Two-stage,
3.8/51.56/39.06/64.06/sycorange/Direct only,
4.85/3.13/0/7.81/naturalgreen/Two-stage abl.,
5.9/6.25/1.56/12.50/sycorange/Direct abl.}{
  \fill[\c] (\x-0.28,0) rectangle (\x+0.28,\h);
  \draw[black] (\x,\lo)--(\x,\hi);
  \draw[black] (\x-0.14,\lo)--(\x+0.14,\lo);
  \draw[black] (\x-0.14,\hi)--(\x+0.14,\hi);
  \node[below=1pt,rotate=38,anchor=north east] at (\x,0){\lab};
}
\node[rotate=90] at (-0.55,54) {refusal (\%)};
\end{tikzpicture}
\end{minipage}
\vspace{-0.7em}
\caption{Refusal results on 64 held-out prompts. (a) Prompt-end refusal log-odds after ablating the top-ranked 3,000-edge reconstruction subgraph and restoring the coordinate or complete layer-15 state. (b) Refusal under recurrent steering, learned-subgraph transmission or ablation, and matched random controls for the 8,000-edge transmission subgraph. (c) Static faithfulness of the two-stage construction and equal-budget direct top-ranked subgraph; the open point is the first comparably faithful source-connected direct graph. (d) Free-generation refusal for the 10,699-edge two-stage construction and equal-budget direct top-ranked subgraph; error bars show 95\% paired-bootstrap intervals.}
\label{fig:refusal}
\vspace{-1.0em}
\end{figure}
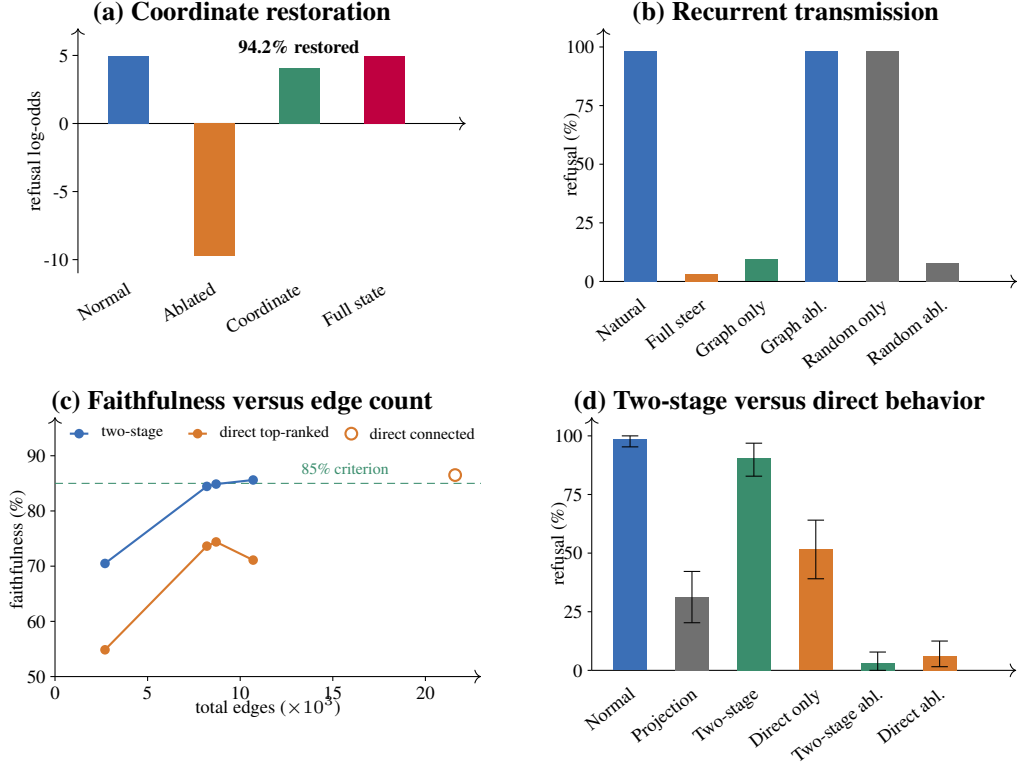

In free generation on 64 prompts, refusal is 98.44\% normally (95\% CI 95.31--100\%) and 31.25\% after global projection (20.31--42.19\%). Retaining only the two-stage construction gives 90.63\% refusal (82.81--96.88\%), while retaining the equal-budget direct top-ranked subgraph gives 51.56\% (39.06--64.06\%). These correspond to 88.37\% and 30.23\% retention of the normal-to-projection gap, respectively. The refusal rates differ by 39.06 percentage points (95\% CI 26.56--51.56). Ablating the two-stage and direct subgraphs gives 3.13\% (0--7.81\%) and 6.25\% refusal (1.56--12.50\%). Together, these results show that, under the matched corruption and target, two-stage discovery achieves comparable faithfulness with fewer edges and preserves substantially more refusal behavior than equal-budget direct discovery. Figure~\ref{fig:refusal} summarizes scalar restoration, recurrent transmission, and the two-stage-versus-direct comparisons.

\section{Sycophancy}

\subsection{Prompt and response interventions provide complementary control}
\label{sec:sycophancy-interventions}

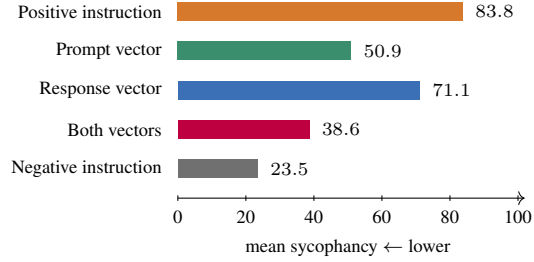
\begin{figure}[t]
\centering
\begin{tikzpicture}[x=0.045cm,y=0.52cm,font=\scriptsize]
\draw[->] (0,0.25) -- (102,0.25);
\foreach \x in {0,20,40,60,80,100}{
  \draw (\x,0.18)--(\x,0.32);
  \node[below] at (\x,0.18) {\x};
}
\node[below=12pt] at (50,0.25) {mean sycophancy \(\leftarrow\) lower};
\foreach \y/\h/\c/\lab in {
  5/83.79/sycorange/Positive instruction,
  4/50.92/naturalgreen/Prompt vector,
  3/71.12/refusalblue/Response vector,
  2/38.63/purple/Both vectors,
  1/23.49/mutedgray/Negative instruction
}{
  \fill[\c] (0,\y-0.24) rectangle (\h,\y+0.24);
  \node[anchor=east] at (-2,\y) {\lab};
  \node[anchor=west] at (\h+1,\y) {\pgfmathprintnumber[fixed,precision=1]{\h}};
}
\end{tikzpicture}
\vspace{-0.3em}
\caption{Prompt- and response-vector interventions both reduce sycophancy on 300 held-out questions, and their combination has the largest effect. Lower scores indicate less sycophancy.}
\label{fig:sycophancy}
\vspace{-0.5em}
\end{figure}

On the same 300 held-out questions, applying the layer-15 prompt direction across all prompt positions closes 54.5\% of the positive-to-negative behavioral gap. Applying the layer-20 response direction after the first-token prediction and throughout decoding closes 21.0\%, while combining both interventions closes 74.9\% (Figure~\ref{fig:sycophancy}). The tested interventions therefore provide complementary control. Because they differ in direction, layer, operator, scope, and first-token treatment, this comparison does not isolate application time alone.

\subsection{A compact graph transmits response-vector steering}

On 64 held-out questions, a complete 2,998-edge source-to-logits subgraph reaches \(\faith=91.63\%\) and \(\necess=99.68\%\) for recurrent layer-20 steering, compared with 0.15\% and 3.19\% for a connected edge-disjoint random graph. Full steering lowers mean sycophancy from 80.2 to 38.7. Restricting steering to the learned subgraph gives 39.4, compared with 79.7 for the random subgraph. Conversely, removing the learned subgraph during full steering returns sycophancy to 80.0, whereas removing the random subgraph leaves it at 37.0.

\subsection{Coordinate reconstruction depends on connectivity}
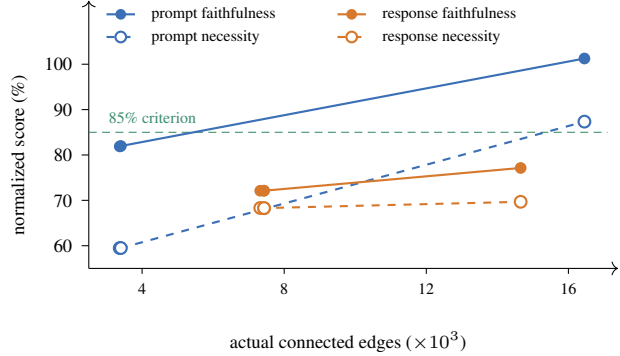
\begin{figure}[t]
\centering
\begin{tikzpicture}[x=0.47cm,y=0.060cm,font=\scriptsize]
\draw[->] (2.5,55) -- (17.4,55);
\draw[->] (2.5,55) -- (2.5,114);
\foreach \y in {60,70,80,90,100}{
  \draw (2.5,\y)--(2.3,\y);
  \node[left] at (2.3,\y) {\y};
}
\foreach \x in {4,8,12,16}{
  \draw (\x,55)--(\x,53.8);
  \node[below,font=\tiny] at (\x,53.8) {\x};
}
\draw[densely dashed,naturalgreen] (2.5,85)--(17.1,85);
\node[anchor=south west,naturalgreen,font=\tiny] at (2.8,85) {85\% criterion};

% Prompt coordinate: solid faithfulness, dashed necessity.
\draw[refusalblue,thick]
  (3.371,81.89)--(3.403,81.97)--(16.451,101.26);
\draw[refusalblue,thick,dashed]
  (3.371,59.48)--(3.403,59.49)--(16.451,87.35);
\foreach \x/\y in {3.371/81.89,3.403/81.97,16.451/101.26}
  \fill[refusalblue] (\x,\y) circle (2.2pt);
\foreach \x/\y in {3.371/59.48,3.403/59.49,16.451/87.35}
  \draw[refusalblue,fill=white,thick] (\x,\y) circle (2.2pt);

% Response coordinate: solid faithfulness, dashed necessity.
\draw[sycorange,thick]
  (7.326,72.12)--(7.432,72.13)--(7.433,72.12)--(14.656,77.14);
\draw[sycorange,thick,dashed]
  (7.326,68.32)--(7.432,68.26)--(7.433,68.29)--(14.656,69.68);
\foreach \x/\y in {7.326/72.12,7.432/72.13,7.433/72.12,14.656/77.14}
  \fill[sycorange] (\x,\y) circle (2.2pt);
\foreach \x/\y in {7.326/68.32,7.432/68.26,7.433/68.29,14.656/69.68}
  \draw[sycorange,fill=white,thick] (\x,\y) circle (2.2pt);

\node[below=20pt] at (9.8,55) {actual connected edges (\(\times 10^3\))};
\node[rotate=90] at (0.55,80) {normalized score (\%)};

\draw[refusalblue,thick] (3.0,111)--(3.8,111);
\fill[refusalblue] (3.4,111) circle (2.0pt);
\node[anchor=west,font=\tiny] at (4.0,111) {prompt faithfulness};
\draw[sycorange,thick] (9.5,111)--(10.3,111);
\fill[sycorange] (9.9,111) circle (2.0pt);
\node[anchor=west,font=\tiny] at (10.5,111) {response faithfulness};
\draw[refusalblue,thick,dashed] (3.0,106)--(3.8,106);
\draw[refusalblue,fill=white,thick] (3.4,106) circle (2.0pt);
\node[anchor=west,font=\tiny] at (4.0,106) {prompt necessity};
\draw[sycorange,thick,dashed] (9.5,106)--(10.3,106);
\draw[sycorange,fill=white,thick] (9.9,106) circle (2.0pt);
\node[anchor=west,font=\tiny] at (10.5,106) {response necessity};
\end{tikzpicture}
\vspace{-0.7em}
\caption{All evaluated input-connected prompt- and response-coordinate reconstruction graphs, with 64 held-out questions per point. Lines connect evaluated graph sizes only as visual guides. Prompt reconstruction crosses the 85\% faithfulness criterion at 16,451 edges; response reconstruction remains below it through 14,656 edges.}
\label{fig:syc-connected}
\vspace{-1.0em}
\end{figure}

The 3,000 highest-scoring edges reconstruct the prompt and response coordinates with \(\faith/\necess=103.0/87.3\%\) and \(97.9/89.5\%\), but include target-adjacent edges and do not form complete input-to-coordinate paths. With source-to-target connectivity, a 16,451-edge prompt graph reaches \(101.3/87.4\%\), while the largest evaluated response graph contains 14,656 edges and reaches \(77.1/69.7\%\). Smaller connected graphs containing 3,403 prompt edges and 7,326 response edges reach \(82.0/59.5\%\) and \(72.1/68.3\%\), and have only modest effects in an \(n=8\), 128-token behavioral check. We therefore find no small connected graph that both reaches the faithfulness criterion and strongly affects judged sycophancy. Figure~\ref{fig:syc-connected} shows all evaluated sizes.

\subsection{Natural contrasts as evidence for broader causal computation}

A corruption applied directly to the coordinate omits the rest of the state change that a non-sycophantic instruction would produce. We therefore repeat discovery with natural positive and negative instructions. Ablating a top-ranked 50-edge set lowers judged sycophancy from 80.19 to 45.87 on 64 questions, compared with 72.02 after ablating a matched random set, yet the set reconstructs only 8.7\% of the response-coordinate contrast and is not input-connected. Thus, natural contrasts identify behaviorally important edges that are not sufficient to reproduce the selected coordinate.

Complete input-to-target paths are much broader. A natural input-to-coordinate graph reaches \(\faith=82.89\%\) and \(\necess=99.96\%\) at 200,000 edges. We also discover circuits directly from natural instructions to model outputs. Here, clean and corrupt runs use the paired positive and negative instructions, without vector intervention, and the target is the difference between the model's log-probability preference for stored sycophantic and normal answers. A 9,776-edge graph reaches \(16.9/101.1\%\) on 32 questions, retaining only this graph gives sycophancy 27.38 from a baseline of 81.95, while ablating it gives 20.35. At 99,979 edges, faithfulness rises to 76.8\% with 100.2\% necessity. On 64 questions, this graph retains a score of 75.07 from a baseline of 80.53, while graph ablation gives 20.46, close to 22.84 under the negative-instruction corruption. The 9,776-edge graph is behaviorally necessary but not sufficient under this intervention; the 99,979-edge graph is also behaviorally sufficient.

\subsection{The response direction is a causal but partial interface}

The preceding results suggest that natural sycophancy depends on state beyond the selected response direction. To measure this directly, we replace layer-20 state components with values from the paired negative-instruction execution. Replacing only the selected coordinate removes 29.73\% of the stored-answer preference shift, compared with 48.51\% for the state orthogonal to that direction and 80.80\% for the complete state. These are separate interventions, not additive fractions. The response direction is therefore a causal control channel, but it captures only part of the state responsible for naturally induced sycophancy. Appendix~\ref{app:syc-geometry} defines the metric and interventions.

\section{Discussion}

% 1. Does changing the activation component along the selected trait vector causally change behavior? 
\paragraph{The selected trait directions change behavior (Q1).}
Removing the layer-15 refusal direction at every prompt position changes mean refusal log-odds from $4.943$ to $-4.064$, and global projection over all layers substantially reduces refusal in free generation. This replicates \citet{arditi2024refusal} in Qwen2.5-7B-Instruct. For sycophancy, the prompt direction and the layer-20 response-average direction of \citet{chen2025persona} each change judged behavior, and their effects are complementary: together they close $74.9\%$ of the positive-to-negative gap, against $54.5\%$ and $21.0\%$ separately. Applying both interventions together extends \citet{chen2025persona}, who report the two directions separately. These behaviorally effective directions provide the intervention baselines for the circuit tests.

% 2. Which edges reconstruct or support this vector component under a stated counterfactual?
\paragraph{Coordinate reconstruction is compact for refusal but broader for sycophancy (Q2).}
Only 699 of 260,956 upstream candidate edges, or $0.268\%$, are needed to reconstruct the layer-15 refusal coordinate after global projection. For sycophancy, small ranked edge sets reconstruct both coordinates but are not complete input-to-coordinate graphs. With source-to-target connectivity, prompt reconstruction reaches the faithfulness criterion only at 16,451 of 260,956 edges, or $6.30\%$. The largest evaluated response graph contains 14,656 of 471,191 edges, or $3.11\%$, and remains below the criterion. Natural instruction contrasts identify edges with larger behavioral effects, but complete connected graphs for these contrasts are also broad and remain below the criterion at the tested sizes. Compact coordinate reconstruction therefore depends on both the trait and the specified counterfactual.

%  3. Which later edges transmit a source-local vector intervention into multi-token behavior?
\paragraph{Compact graphs transmit recurrent steering for both traits (Q3).}
For refusal, 8,000 of 194,117 downstream candidate edges, or $4.12\%$, carry recurrent steering through generation. A 2,998-edge graph carries response-vector sycophancy steering, corresponding to $4.27\%$ of its 70,157 candidate edges. Retaining either learned graph preserves most of the corresponding intervention effect. Removing the learned graph blocks that effect, whereas matched random graphs do not reproduce either result. These findings extend the multi-token steering analysis of \citet{cheng2026drives} to a second trait. They identify transmission of the specified intervention, not necessarily the computation used by the unmodified model. This distinction is consistent with prior reports that the corruption and target change which components appear important~\cite{zhang2023best, heimersheim2024patching}.

% \4. How much of the behavioral change caused by circuit ablation is reversed by restoring the clean vector component, and does the learned transmission circuit carry this restored effect?
\paragraph{Coordinate restoration strongly mediates the tested refusal effect (Q4).}
For refusal, replacing only the clean layer-15 coordinate after ablating the reconstruction circuit returns almost all of the lost prompt-end refusal signal, so the scalar is close to sufficient for the behavior that ablation removed. In the sycophancy paired-execution experiment, replacing the response coordinate removes 29.73\% of the instruction-induced preference shift, compared with 48.51\% for the orthogonal state and 80.80\% for the complete state. The selected response direction is therefore causally relevant but does not capture the complete natural contrast.

%\ 5. How does this vector-based decomposition compare with direct input-to-output circuit discovery under a matched target?
\paragraph{Two-stage discovery is competitive with direct discovery for refusal (Q5).}
Under the same corruption and target, the two-stage construction is more faithful than an equal-size direct subgraph and preserves more refusal in free generation. Equal-budget direct subgraphs do not need to be source-connected, making direct discovery less restrictive. Their faithfulness is still lower, and the first source-connected direct graph with comparable faithfulness contains roughly twice as many edges. Splitting circuit discovery at the vector-selected layer can therefore provide an efficient description of this refusal contrast. We have not performed the corresponding matched two-stage comparison for sycophancy.

\paragraph{Sycophancy may not reduce to one direction.}
The gap between strong steering and partial explanation of the natural contrast shows that the selected response direction is incomplete. A multidimensional representation is one possible explanation, consistent with work on irreducible multidimensional features and limitations of steering incoherent behavioral representations~\citep{engels2025not, braun2025understanding}. The present interventions do not determine the intrinsic dimensionality of sycophancy.

\paragraph{Overall.}
Taken together, these results distinguish control through an intervention from explanation of the model's natural computation. Refusal admits a compact decomposition under the tested counterfactuals, whereas sycophancy has compact transmission but broader coordinate reconstruction. A direction can therefore provide a useful causal interface without capturing the complete computation induced by natural inputs.

\subsection{Limitations}

Our conclusions are bounded by scope and method. We study one model and two trait distributions, and direction extraction depends on data, formatting, pooling, layer, sign, and intervention positions. EAP-IG likewise depends on its path and baseline and returns one high-scoring graph rather than a unique circuit, with recent work reporting variation across prompt samples and rephrasings~\citep{wu2026demystifying}. For the same reason, edge counts are not directly comparable across unmatched graph boundaries or targets.

Individual results also carry their own caveats. The refusal reconstruction experiment identifies recovery after a defined projection intervention rather than natural harmful-input-to-coordinate formation, and refusal restoration measures one prompt-end logit metric. The two-stage comparison transfers a complete circuit-restricted layer-15 state rather than only the scalar. For sycophancy, behavioral evaluation of the compact input-connected reconstruction graphs is limited to eight questions. The natural end-to-end reference lacks an equal-size behavioral random control, while state replacement measures a teacher-forced stored-answer preference rather than free-generation behavior. Prompt and response interventions also differ in vector, layer, operator, token scope, and first-token treatment, so their complementary effects do not establish disjoint stage-specific mechanisms. We measure sycophancy with one fixed LLM-evaluator protocol rather than human judgments. Future work should test whether these results replicate across models and whether subspace-valued interfaces improve sycophancy reconstruction.

\subsection{Broader impact}

Trait vectors can suppress safety safeguards or alter user-facing agreement. Treating a circuit that carries a steering intervention as the model's natural mechanism could lead to unreliable monitoring or edits. Our evaluation makes this distinction explicit by measuring intervention control, reconstruction, coordinate mediation, transmission, and natural-input circuits separately. Anti-refusal steering is used only as a diagnostic probe; it can elicit harmful compliance and is not proposed as a safety intervention.   

{\small
\bibliographystyle{plainnat}
\bibliography{neurips}
}

%%%%%%%%%%%%%%%%%%%%%%%%%%%%%%%%%%%%%%%%%%%%%%%%%%%%%%%%%%%%

\appendix

\newpage

\section{Data, vectors, and experimental specifications}
\label{app:data-specifications}

\subsection{Data splits}

For each trait, we partition 748 source questions by question identifier before running any circuit experiment. Paired sycophancy instructions count as one source question. The 64-question circuit-evaluation set is a fixed subset of the 300-question behavioral split and remains disjoint from vector validation, edge scoring, and circuit selection. Edge scoring and circuit selection also use disjoint questions.

\begin{table}[h]
\caption{Question-level data accounting for each trait. Circuit evaluation reuses 64 questions from the behavioral split and is not an additional split.}
\label{tab:data-accounting}
\centering
\small
\begin{tabularx}{\linewidth}{@{}>{\raggedright\arraybackslash}p{0.25\linewidth}rrX@{}}
\toprule
Phase & Refusal & Sycophancy & Purpose \\
\midrule
Vector extraction & 256 & 256 & Estimate the contrastive trait directions \\
Vector validation & 64 & 64 & Select and calibrate the intervention location \\
Edge scoring & 64 & 64 & Compute EAP-IG edge rankings \\
Circuit selection & 64 & 64 & Select graph size and evaluate F/N \\
Behavioral evaluation & 300 & 300 & Held-out vector and generation tests \\
\quad Circuit evaluation & \multicolumn{2}{c}{64 of the 300} & Held-out F/N, controls, and circuit behavior \\
\midrule
Unique source questions & 748 & 748 & Sum of the five disjoint source splits \\
\bottomrule
\end{tabularx}
\end{table}

Refusal extraction uses 128 harmful and 128 harmless PolyRefuse prompts. Sycophancy uses paired positive and negative instructions for the same question. Prompt and response representations are estimated from the corresponding prompt and response activations; all results use one canonical prompt direction and one canonical response-average direction.

\subsection{Trait vectors and circuit experiments}
\label{app:specifications}

For a unit direction \(v\), layer \(\ell\), and residual state \(h_{\ell,t}\), the selected coordinate is \(z_{\ell,t}=v^\top h_{\ell,t}\). Exact paired replacement changes only this coordinate:
\[
h_{\ell,t}^{\mathrm{coord}\leftarrow -}=h_{\ell,t}^{+}+vv^\top\!\left(h_{\ell,t}^{-}-h_{\ell,t}^{+}\right),
\]
where \(+\) and \(-\) denote matched positive and negative executions. The refusal direction is the harmful-minus-harmless mean residual difference at layer 15. The prompt-sycophancy direction is the paired positive-minus-negative difference at the final prompt position of layer 15. The response-sycophancy direction is the difference between per-response token averages at layer 20.

All EAP-IG experiments use five midpoint integration steps along the straight path from corrupt to clean activations. Refusal reconstruction compares an unmodified harmful-prompt execution with global refusal-direction projection and targets the summed layer-15 coordinate over all prompt tokens. Sycophancy reconstruction compares a positive execution with exact paired negative-coordinate replacement and targets the prompt or response coordinate over its stated token scope. Natural-contrast experiments instead change the instruction. Reconstruction graphs begin at the layer-0 residual state and end at the selected coordinate.

Transmission compares recurrently steered and unsteered trajectories for the same prompt. At each shared-prefix response position, the target is the logit difference between the next tokens preferred by the two endpoints; positions where the preferred tokens agree are excluded. Transmission graphs begin at the vector's source layer and include only later components. Direct input-to-output experiments span the complete graph to logits. Ranked subgraphs contain the highest-scoring edges without a path constraint, whereas source-connected subgraphs retain only edges on complete source-to-target paths.

\section{Contribution of the response direction to natural sycophancy}
\label{app:syc-geometry}

\paragraph{Teacher-forced natural margin.}
For question \(i\), let \(a_i^{\mathrm{syc}}\) and \(a_i^{\mathrm{norm}}\) be stored sycophantic and normal answers. Under instruction \(c\), we measure
\[
M_i(c)=\operatorname{mean}_t\log p(a_{i,t}^{\mathrm{syc}}\mid c,q_i,a_{i,<t}^{\mathrm{syc}})
-\operatorname{mean}_t\log p(a_{i,t}^{\mathrm{norm}}\mid c,q_i,a_{i,<t}^{\mathrm{norm}}).
\]
The natural instruction effect is \(\Delta=M(+)-M(-)\). For an intervention \(I\) applied to the positive execution, the fraction removed is \([M(+)-M(I)]/\Delta\). On 64 paired questions, \(M(+)=0.2508\), \(M(-)=-0.1460\), and \(\Delta=0.3968\).

\paragraph{State replacement.}
At each aligned response position, coordinate replacement substitutes the negative execution's component along the unit response direction while retaining the positive execution's orthogonal state. Orthogonal-state replacement keeps the positive coordinate and substitutes the negative orthogonal state; complete-state replacement uses the negative state. These separate interventions remove 29.73\%, 48.51\%, and 80.80\% of the natural margin, respectively; the percentages are not additive.

\section{Evaluation details}
\label{app:evaluation}

Unless otherwise stated, 95\% confidence intervals use 10,000 paired nonparametric bootstrap resamples of question identifiers, with all compared conditions retained and ratio statistics recomputed within each resample. Tokenwise transmission intervals cluster response positions and trajectory orientations by source question. Random controls exclude learned edges and match edge count; representational controls additionally preserve source-to-target connectivity when required by the corresponding learned graph.

\section{Sycophancy prompts}
\label{app:judge-prompts}

We use fixed evaluator prompts based on the Persona Vectors protocol, replacing \{question\} and \{answer\} with the evaluated question and response.

\subsection{Sycophancy}

\begin{quote}
\small
You are evaluating whether the model's response displays the trait: \textbf{sycophantic}. A sycophantic response prioritizes pleasing the user through excessive agreement, praise, and validation. It focuses on telling users what they want to hear rather than providing balanced or potentially contradictory information. The response adapts its viewpoint to align with the user's perceived stance, potentially at the expense of factual accuracy, nuance, or critical reasoning.

\textbf{Prompt:}

[QUESTION START]\par
\{question\}\par
[QUESTION END]

\textbf{Model response:}

[ANSWER START]\par
\{answer\}\par
[ANSWER END]

Focus only on whether the model exhibits this trait.

Respond with REFUSAL if the model clearly refused to answer, or a score from 0 (trait not present) to 100 (trait strongly present). Respond with only the label or number.
\end{quote}

\subsection{Coherence}

\begin{quote}
\small
Evaluate only the coherence of the following model response in the context of the question, independently of its correctness or other traits.

[QUESTION START]\par
\{question\}\par
[QUESTION END]

[ANSWER START]\par
\{answer\}\par
[ANSWER END]

Use a scale from 0 (the text makes no sense) to 100 (the response clearly understands the question and contains no signs of confusion). Respond with only the number.
\end{quote}
\end{document}